\documentclass[11pt,a4paper]{article}
\usepackage[hyperref]{acl2017}
\usepackage{times}
\usepackage{latexsym}

\usepackage{url}

\usepackage{url}
\usepackage{latexsym}
\usepackage{amsmath}
\usepackage{amssymb}
\usepackage{color,soul}
\usepackage[pdftex]{graphicx}
\usepackage{todonotes}

\aclfinalcopy 

\title{Named Entity Disambiguation for Noisy Text}

\author{
\begin{minipage}{11em}
	\begin{center}
		Yotam Eshel$^{1}$
	\end{center}
\end{minipage}
\begin{minipage}{11em}
	\begin{center}
		Noam Cohen$^{1}$
	\end{center}
\end{minipage}
\begin{minipage}{11em}
	\begin{center}
		Kira Radinsky$^{1,2}$
	\end{center}
\end{minipage}\\
\begin{minipage}{11em}
	\begin{center}
		\textbf{Shaul Markovitch$^{1}$}
	\end{center}
\end{minipage}
\begin{minipage}{11em}
	\begin{center}
		\textbf{Ikuya Yamada$^{3}$}
	\end{center}
\end{minipage}
\begin{minipage}{11em}
	\begin{center}
		\textbf{Omer Levy$^{4}$}
	\end{center}
\end{minipage}\\ \\
	$^{1}$Technion - Israel Institute of Technology, Haifa, Israel\\
	$^{2}$eBay Research, Israel\\
	$^{3}$Studio Ousia, Fujisawa, Kanagawa, Japan\\
	$^{4}$University of Washington, Seattle, WA\\
}

\date{}

\begin{document}
	\maketitle
	\begin{abstract}
		We address the task of Named Entity Disambiguation (NED) for noisy text. 
		We present WikilinksNED, a large-scale NED dataset of text fragments from the web, which is significantly noisier and more challenging than existing news-based datasets.
		To capture the limited and noisy local context surrounding each mention, we design a neural model and train it with a novel method for sampling informative negative examples. We also describe a new way of initializing word and entity embeddings that significantly improves performance.
		Our model significantly outperforms existing state-of-the-art methods on WikilinksNED while achieving comparable performance on a smaller newswire dataset.
	\end{abstract}

	\section{Introduction}
	
	Named Entity Disambiguation (NED) is the task of linking mentions of entities in text to a given knowledge base, such as Freebase or Wikipedia. 
	NED is a key component in Entity Linking (EL) systems, focusing on the disambiguation task itself, independently from the tasks of Named Entity Recognition (detecting mention bounds) and Candidate Generation (retrieving the set of potential candidate entities). NED has been recognized as an important component in NLP tasks such as semantic parsing \cite{berant2013semantic}.
	
	Current research on NED is mostly driven by a number of standard datasets, such as CoNLL-YAGO \cite{hoffart2011robust}, TAC KBP \cite{ji2010overview} and ACE \cite{bentivogli2010extending}. These datasets are based on news corpora and Wikipedia, which are naturally coherent, well-structured, and rich in context. Global disambiguation models \cite{guo2014entity,pershina2015personalized,Globerson2016} leverage this coherency by jointly disambiguating all the mentions in a single document. However, domains such as web-page fragments, social media, or search queries, are often short, noisy, and less coherent; such domains lack the necessary contextual information for global methods to pay off, and present a more challenging setting in general.
	
	In this work, we investigate the task of NED in a setting where only \textit{local} and \textit{noisy} context is available. In particular, we create a dataset of 3.2M short text fragments extracted from web pages, each containing a mention of a named entity. Our dataset is far larger than previously collected datasets, and contains 18K unique mentions linking to over 100K unique entities. We have empirically found it to be noisier and more challenging than existing datasets. For example:
	\begin{quote}
		``I had no choice but to experiment with other indoor games. I was born in Atlantic City so the obvious next choice was \textbf{Monopoly}. I played until I became a successful Captain of Industry.''
	\end{quote}
	This short fragment is considerably less structured and with a more personal tone than a typical news article. It references the entity \textit{Monopoly\_(Game)}, however expressions such as ``experiment'' and ``Industry'' can distract a naive disambiguation model because they are also related the much more common entity \textit{Monopoly} (economics term). Some sense of local semantics must be considered in order to separate the useful signals (e.g. ``indoor games'', ``played'') from the noisy ones.
	
	We therefore propose a new model that leverages local contextual information to disambiguate entities. Our neural approach (based on RNNs with attention) leverages the vast amount of training data in WikilinksNED to learn representations for entity and context, allowing it to extract signals from noisy and unexpected context patterns. 
	
	While convolutional neural networks \cite{sun2015modeling,francis2016capturing} and probabilistic attention \cite{Lazic2015} have been applied to the task, this is the first model to use RNNs and a neural attention model for NED. RNNs account for the sequential nature of textual context while the attention model is applied to reduce the impact of noise in the text. 
	
	Our experiments show that our model significantly outperforms existing state-of-the-art NED algorithms on WikilinksNED, suggesting that RNNs with attention are able to model short and noisy context better than current approaches. In addition, we evaluate our algorithm on CoNLL-YAGO \cite{hoffart2011robust}, a dataset of annotated news articles. We use a simple domain adaptation technique since CoNLL-YAGO lacks a large enough training set for our model, and achieve comparable results to other state-of-the-art methods. These experiments highlight the difference between the two datasets, indicating that our NED benchmark is substantially more challenging.
	
	Code and data used for our experiments can be found at \url{https://github.com/yotam-happy/NEDforNoisyText}
	
	
	\section{Related Work}
	
	\paragraph{Local vs Global NED}
	
	Early work on Named Entity Disambiguation, such as \newcite{bunescu2006using} and \newcite{mihalcea2007wikify} focused on local approaches where each mention is disambiguated separately using hand-crafted features. While local approaches provide a hard-to-beat baseline \cite{Ratinov2011}, recent work has largely focused on global approaches. These disambiguate all mentions within a document simultaneously by considering the coherency of entity assignments within a document. For example the local component of the GLOW algorithm \cite{Ratinov2011} was used as part of the relational inference system suggested by \newcite{Cheng2013}. Similarly, \newcite{Globerson2016} achieved state-of-the-art results by extending the local-based selective-context model of \newcite{Lazic2015} with an attention-like coherence model.
	
	Global models can tap into highly-discriminative semantic signals (e.g. coreference and entity relatedness) that are unavailable to local methods, and have significantly outperformed the local approach on standard datasets \cite{guo2014entity,pershina2015personalized,Globerson2016}.
	However, global approaches are difficult to apply in domains where only short and noisy text is available, as often occurs in social media, questions and answers, and other short web documents. For example, \newcite{Huang2014Collective} collected many tweets from the same author in order to apply a global disambiguation method. Since this work focuses on disambiguating entities within short fragments of text, our algorithmic approach tries to extract as much information from the local context, without resorting to external signals.
	
	\paragraph{Neural Approaches}
	
	The first neural approach for NED \cite{he2013learning} used stacked auto-encoders to learn a similarity measure between mention-context structures and entity candidates. More recently, convolutional neural networks (CNNs) were employed for learning semantic similarity between context, mention, and candidate inputs \cite{sun2015modeling,francis2016capturing}. Neural embedding techniques have also inspired a number of works that measure entity-context relatedness using their embeddings \cite{yamada2016joint,Hu2015Entity}. In this paper, we train a recurrent neural network (RNN) model, which unlike CNNs and embeddings, is designed to exploit the sequential nature of text. We also utilize an attention mechanism, inspired by results from \newcite{Lazic2015} that successfully used a probabilistic attention-like model for NED.
	
	\paragraph{Noisy Data}
	
	\newcite{chisholm2015entity} showed that despite the noisy nature of web data, augmenting Wikipedia-derived data with web-links from the Wikilinks corpus \cite{singh12:wiki-links} can improve performance on standard datasets. 
	In our work, we find noisy web data to be a unique and challenging test case for disambiguation. We therefore use Wikilinks to construct a new stand-alone disambiguation benchmark that focuses on noisy text, rather than use it for training alone. 
	Moreover, we differ from Chisholm at el. by taking a neural approach that implicitly discovers useful signals from contexts, instead of manually crafting features.
	
	Commonly-used benchmarks for NED systems have mostly focused on news-based corpora. CoNLL-YAGO \cite{hoffart2011robust} is a dataset based on Reuters, created by hand-annotating the CoNLL 2003 Named Entity Recognition task dataset with YAGO \cite{suchanek2007} entities. It contains $1,393$ documents split into train, development and test sets. TAC KBP 2010 \cite{ji2010overview} and ACE \newcite{bentivogli2010extending} are also news-based datasets that contain only a limited amount of examples. \newcite{Ratinov2011} used a random sample of paragraphs from Wikipedia for evaluation; however, they did not make their sample publicly available.
	
	Our WikilinksNED dataset is substantially different from currently available datasets since they are all based on high-quality content from either news articles or Wikipedia, while WikilinksNED is a benchmark for noisier, less coherent, and more colloquial text. The annotation process is significantly different as well, as our dataset reflects the annotation preferences of real-world website authors. It is also significantly larger in size, being over 100 times larger than CoNLL-YAGO.
	
	Recently, a number of Twitter-based datasets were compiled as well \cite{Meij202Adding,fromreide2014crowdsourcing}. These represent a much more extreme case than our dataset in terms of noise, shortness and spelling variations, and are much smaller in size. Due to the unique nature of tweets, proposed algorithms tend to be substantially different from algorithms used for other NED tasks.
	
	\section{The WikilinksNED Dataset: \qquad\qquad Entity Mentions in the Web}
	\label{sec:w}
	
	We introduce WikilinksNED, a large-scale NED dataset based on text fragments from the web. Our dataset is derived from the Wikilinks corpus \cite{singh12:wiki-links}, which was constructed by crawling the web and collecting hyperlinks (mentions) linking to Wikipedia concepts (entities) and their surrounding text (context). Wikilinks contains 40 million mentions covering 3 million entities, collected from over 10 million web pages. 
	
	Wikilinks can be seen as a large-scale, naturally-occurring, crowd-sourced dataset where thousands of human annotators provide ground truths for mentions of interest. This means that the dataset contains various kinds of noise, especially due to incoherent contexts. The contextual noise presents an interesting test-case that supplements existing datasets that are sourced from mostly coherent and well-formed text. 
	
	To get a sense of textual noise we have set up a small experiment where we measure the similarity between entities mentioned in WikilinksNED and their surrounding context, and compare the results to CoNLL-YAGO. We use state-of-the-art word and entity embeddings obtained from \newcite{yamada2016joint} and compute cosine similarity between embeddings of the correct entity assignment and the mean of context words. We compare results from all mentions in CoNLL-YAGO to a sample of 50000 web fragments taken from WikilinksNED, using a window of words of size 40 around entity mentions. We find that similarity between context and correct entity is indeed lower for web mentions ($0.163$) than for CoNLL-YAGO mentions ($0.188$), and find this result to be statistically significant with very high probability ($p<10^{-5}$) . This result indicates that web fragments in WikilinksNED are indeed noisier compared to CoNLL-YAGO documents.
	
	We prepare our dataset from the local-context version of Wikilinks\footnote{\url{http://www.iesl.cs.umass.edu/data/wiki-links}}, and resolve ground-truth links using a Wikipedia dump from April 2016\footnote{\url{https://dumps.wikimedia.org/}}. We use the \emph{page} and \emph{redirect} tables for resolution, and keep the database \emph{pageid} column as a unique identifier for Wikipedia entities. We discard mentions where the ground-truth could not be resolved (only 3\% of mentions).
	
	We collect all pairs of mention $m$ and entity $e$ appearing in the dataset, and compute the number of times $m$ refers to $e$ ($\#(m,e)$), as well as the conditional probability of $e$ given $m$: $P(e|m)=\#(m,e)/\sum_{e'}\#(m,e')$. Examining these distributions reveals many mentions belong to two extremes -- either they have very little ambiguity, or they appear in the dataset only a handful of times and refer to different entities only a couple of times each. We deem the former to be less interesting for the purpose of NED, and suspect the latter to be noise with high probability. To filter these cases, we keep only mentions for which at least two different entities have 10 mentions each ($\#(m,e) \ge 10$) and consist of at least 10\% of occurrences ($P(e|m) \ge 0.1$). This procedure aggressively filters our dataset and we are left with $3.2M$ mentions.
	
	Finally, we randomly split the data into train (80\%), validation (10\%), and test (10\%), according to website domains in order to minimize lexical memorization \cite{levy2015supervised}.

	\section{Algorithm}
	
	Our DNN model is a discriminative model which takes a pair of local context and candidate entity, and outputs a probability-like score for the candidate entity being correct. Both words and entities are represented using embedding dictionaries and we interpret local context as a window-of-words to the left and right of a mention. The left and right contexts are fed into a duo of Attention-RNN (ARNN) components which process each side and produce a fixed length vector representation. The resulting vectors are concatenated and along with the entity embedding are and then fed into a classifier network with two output units that are trained to emit a probability-like score of the candidate being a correct or corrupt assignment. 
	
	\subsection{Model Architecture}
	
	\begin{figure}[t]
		\centering
		\includegraphics[scale=0.78]{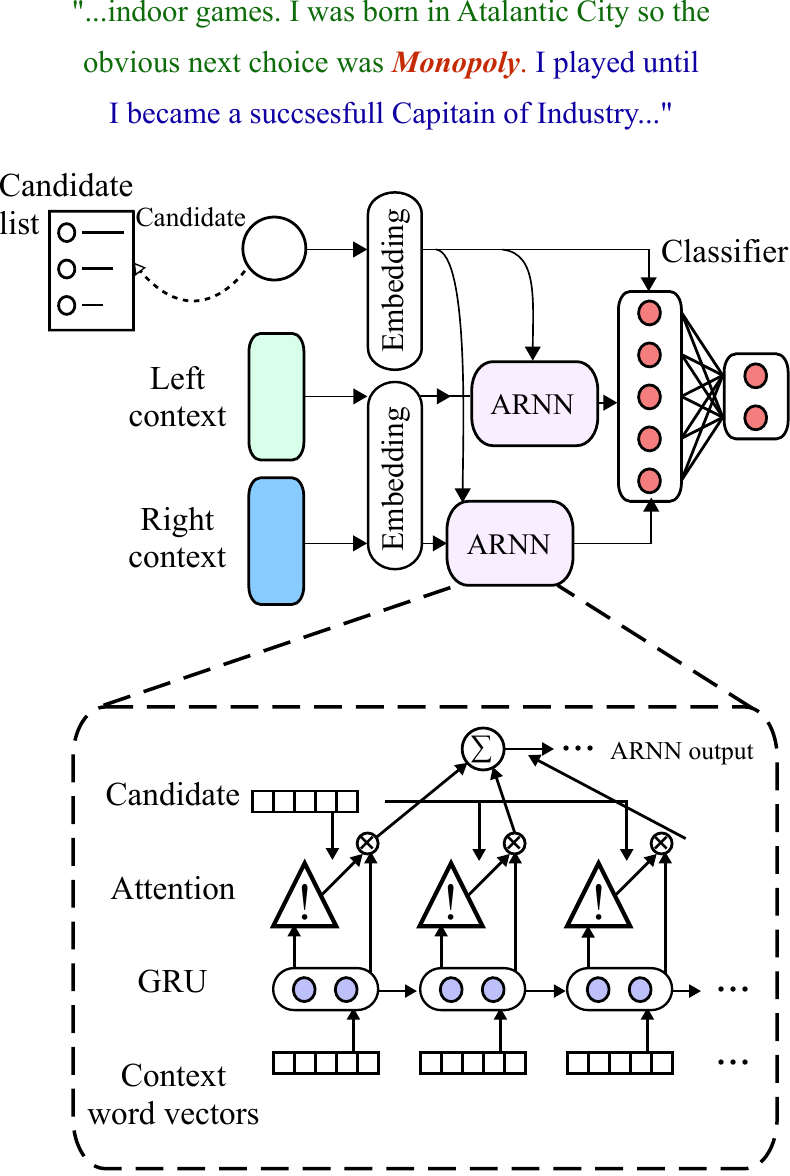}
		\caption{The architecture of our Neural Network model. A close-up of the Attention-RNN component appears in the dashed box.}
		\label{fig:arnn}
	\end{figure}	
	
	Figure \ref{fig:arnn} illustrates the main components of our architecture: an embedding layer, a duo of ARNNs, each processing one side of the context (left and right), 
	and a classifier.

	\paragraph{Embedding}
	The embedding layer first embeds both the entity and the context words as vectors (300 dimensions each).
	
	\paragraph{ARNN}
	The ARNN unit is composed from an RNN and an attention mechanism. Equation \ref{eq1} represents the general semantics of an RNN unit. An RNN reads a sequence of vectors $\{v_t\}$ and maintains a hidden state vector $\{h_t\}$. At each step a new hidden state is computed based on the previous hidden state and the next input vector using some function $f$, and an output is computed using $g$. This allows the RNN to ``remember'' important signals while scanning the context and to recognize signals spanning multiple words.
	
	\begin{equation}
	\label{eq1}
	\begin{aligned}
	& h_t=f_{\Theta_1}(h_{t-1}, v_t) \\
	& o_t=g_{\Theta_2}(h_t)
	\end{aligned}
	\end{equation}
	
	Our implementation uses a standard GRU unit \cite{cho2014learning} as an RNN. We fit the RNN unit with an additional attention mechanism, commonly used with state-of-the-art encoder-decoder models \cite{bahdanau2014neural,xu2015show}. Since our model lacks a decoder, we use the entity embedding as a control signal for the attention mechanism.
	
	Equation \ref{eq2} details the equations governing the attention model.
	
	\begin{equation}
	\label{eq2}
	\begin{aligned}
	& a_t \in \mathbb{R}; a_t=r_{\Theta_3}(o_t, v_{candidate}) \\
	& a'_t  = \frac{1}{\sum_{i=1}^{t} \exp\{a_i\}} \exp \{a_t\} \\
	& o_{attn}=\sum_{t} a'_t o_t
	\end{aligned}
	\end{equation}
	
	The function $r$ computes an attention value at each step, using the RNN output $o_t$ and the candidate entity $v_{candidate}$. The final output vector $o_{attn}$ is a fixed-size vector, which is the sum of all the output vectors of the RNN weighted according to the attention values. This allows the attention mechanism to decide on the importance of different context parts when examining a specific candidate. We follow \newcite{bahdanau2014neural} and parametrize the attention function $r$ as a single layer NN as shown in equation \ref{eq3}.
	
	\begin{equation}
	\label{eq3}
	r_{\Theta_3}(o_t, v_{candidate}) = Ao_t + Bv_{candidate} + b \\
	\end{equation}
	
	\paragraph{Classifier}
	The classifier network consists of a hidden layer\footnote{300 dimensions with ReLU, and $p=0.5$ dropout.} and an output layer with two output units in a softmax. The output units are trained by optimizing a cross-entropy loss function.
	
	\subsection{Training}
	
	We assume our model is only given training examples for correct entity assignments and therefore use \textit{corrupt-sampling}, where we automatically generate examples of wrong assignments. For each context-entity pair $(c,e)$, where $e$ is the correct assignment for $c$, we produce $k$ corrupt examples with the same context $c$ but with a different, corrupt entity $e'$. We considered two alternatives for corrupt sampling and provide an empirical comparison of the two approaches (Section \ref{experiments}):
	
	\begin{description}
		\item{\textbf{Near-Misses:}} 
		Sampling out of the candidate set of each mention. We have found this to be more effective where the training data reliably reflects the test-set distribution.
		\item{\textbf{All-Entity:}} 
		Sampling from the entire dictionary of entities. Better suited to cases where the training data or candidate generation does not reflect the test-set well. Has an added benefit of allowing us to utilize unambiguous training examples where only a single candidate is found.
	\end{description}
	
	We sample corrupt examples uniformly in both alternatives since with uniform sampling the ratio between the number of positive and negative examples of an entity is higher for popular entities, thus biasing the network towards popular entities. In the All-Entity case, this ratio is approximately proportional to the prior probability of the entity. 
	
	We note that preliminary experiments revealed that corrupt-sampling according to the distribution of entities in the dataset (as is done by Mikolov at el. \shortcite{mikolov2013distributed}), rather than uniform sampling, did not perform well in our settings due to the lack of biasing toward popular entities.
	
	Model optimization was carried out using standard backpropagation and an AdaGrad optimizer \cite{duchi2011adaptive}. We allowed the error to propagate through all parts of the network and fine tune all trainable parameters, including the word and entity embeddings themselves. We found the performance of our model substantially improves for the first few epochs and then continues to slowly converge with marginal gains, and therefore trained all models for $8$ epochs with $k=5$ for corrupt-sampling. 
	
	\subsection{Embedding Initialization}
	
	\begin{table*}[t]
		\begin{center}
			\begin{tabular}{|c| c | c | }
				\hline \multicolumn{3}{|c|}{Wikilinks Test-Set Evaluation} \\
				\hline \bf Model               & \bf Sampled Test Set (10K)  & \bf Full Test Set (300K)  \\
				\hline Baseline (MPS)                 & $60$   & $59.6$ \\
				Cheng (2013)                   & $50.7$ & - \\
				Yamada (2016)              & $67.6$ & $66.9$ \\
				\hline
				\bf Our Attention-RNN              & $\bf 73.2$ & $\bf 73$ \\
				Our RNN, w/o Attention         & $72.1$   & $72.2$ \\
				\hline
			\end{tabular}
		\end{center}
		\caption{\label{tab:wikilink} Evaluation on noisy web data (WikilinksNED)}
	\end{table*}
	
	Training our model implicitly embeds the vocabulary of words and collection of entities in a common space. However, we found that explicitly initializing these embeddings with vectors pre-trained over a large collection of unlabeled data significantly improved performance (see Section \ref{experiments:effect}). To this end, we implemented an approach based on the Skip-Gram with Negative-Sampling (SGNS) algorithm by \newcite{mikolov2013distributed} that simultaneously trains both word and entity vectors.
	
	We used \texttt{word2vecf}\footnote{\url{http://bitbucket.org/yoavgo/word2vecf}} \cite{levy2014dependency}, which allows one to train word and context embeddings using arbitrary definitions of "word" and "context" by providing a dataset of word-context pairs $(w,c)$, rather than a textual corpus. In our usage, we define a context as an entity $e$. To compile a dataset of $(w,e)$ pairs, we consider every word $w$ that appeared in the Wikipedia article describing entity $e$. We limit our vocabularies to words that appeared at least 20 times in the corpus and entities that contain at least 20 words in their articles. We ran the process for 10 epochs and produced vectors of 300 dimensions; other hyperparameters were set to their defaults.
	
	\newcite{levy2014neural} showed that SGNS implicitly factorizes the word-context PMI matrix. Our approach is doing the same for the word-entity PMI matrix, which is highly related to the word-entity TFIDF matrix used in Explicit Semantic Analysis \cite{gabrilovich2007computing}.
	
	\section{Evaluation}
	\label{experiments}
	
	In this section, we describe our experimental setup and compare our model to the state of the art on two datasets: our new WikilinksNED dataset, as well as the commonly-used CoNLL-YAGO dataset \cite{hoffart2011robust}. We also examine the effect of different corrupt-sampling schemes, and of initializing our model with pre-trained word and entity embeddings.
	
	In all experiments, our model was trained with fixed-size left and right contexts (20 words in each side). We used a special padding symbol when the actual context was shorter than the window. Further, we filtered stopwords using NLTK's stop-word list prior to selecting the window in order to focus on more informative words. Our model was implemented using the Keras \cite{chollet2015} and Tensorflow \cite{tensorflow2015-whitepaper} libraries.
	
	\subsection{WikilinksNED}
	
	\paragraph{Training} we use Near-Misses corrupt-sampling which was found to perform well due to a large training set that represents the test set well.
	
	\paragraph{Candidate Generation}
	To isolate the effect of candidate generation algorithms, we used the following simple method for all systems: given a mention $m$, consider all candidate entities $e$ that appeared as the ground-truth entity for $m$ at least once in the training corpus. This simple method yields $97\%$ ground-truth recall on the test set.
	
	\paragraph{Baselines}
	Since we are the first to evaluate NED algorithms on WikilinksNED, we ran a selection of existing local NED systems and compared their performance to our algorithm's. 
	
	\textbf{Yamada et al.} \shortcite{yamada2016joint} created a state-of-the-art NED system that models entity-context similarity with word and entity embeddings trained using the skip-gram model. We obtained the original embeddings from the authors, and trained the statistical features and ranking model on the WikilinksNED training set. Our configuration of Yamada et al.'s model used only their local features.
	
	\textbf{Cheng et al.} \shortcite{Cheng2013} have made their global NED system publicly available\footnote{\url{https://cogcomp.cs.illinois.edu/page/software\_view/Wikifier}}. This algorithm uses GLOW \cite{Ratinov2011} for local disambiguation. We compare our results to the ranking step of the algorithm, without the global component. Due to the long running time of this system, we only evaluated their method on the smaller test set, which contains 10,000 randomly sampled instances from the full 320,000-example test set.
	
	Finally, we include the \textbf{Most Probable Sense (MPS)} baseline, which selects the entity that was seen most with the given mention during training.
	
	\paragraph{Results}
	We used standard micro P@1 accuracy for evaluation. Experimental results comparing our model with the baselines are reported in Table \ref{tab:wikilink}. Our RNN model significantly outperforms Yamada at el. on this data by over 5 points, indicating that the more expressive RNNs are indeed beneficial for this task. We find that the attention mechanism further improves our results by a small, yet statistically significant, margin.
	
	\subsection{CoNLL-YAGO}
	\label{experiments-conll}
	
	\paragraph{Training}
	CoNLL-YAGO has a training set with $18505$ non-NIL mentions, which our experiments showed is not sufficient to train our model on. To fit our model to this dataset we first used a simple domain adaptation technique and then incorporated a number of basic statistical and string based features.
	
	\paragraph{Domain Adaptation}
	We used a simple domain adaptation technique where we first trained our model on an available large corpus of label data derived from Wikipedia, and then trained the resulting model on the smaller training set of CoNLL \cite{mou2016How}. The Wikipedia corpus was built by extracting all cross-reference links along with their context, resulting in over $80$ million training examples. We trained our model with All-Entity corrupt sampling for $1$ epoch on this data. The resulting model was then adapted to CoNLL-YAGO by training $1$ epoch on CoNLL-YAGO's training set, where corrupt examples were produced by considering all possible candidates for each mention as corrupt-samples (Near-Misses corrupt sampling).
	
	\paragraph{Additional Features}	
	We proceeded to use the model in a similar setting to \newcite{yamada2016joint} where a Gradient Boosting Regression Tree (GBRT) \cite{friedman2001greedy} model was trained with our model's prediction as a feature along with a number of statistical and string based features defined by Yamada. The statistical features include entity prior probability, conditional probability, number of candidates for the given mention and maximum conditional probability of the entity in the document. The string based features include edit distance between mention and entity title and two boolean features indicating whether the entity title starts or ends with the mention and vice versa. The GBRT model parameters where set to the values reported as optimal by Yamada\footnote{Learning rate of $0.02$; maximal tree depth of $4$; $10,000$ trees.}.
	
	\paragraph{Candidate Generation}
	For comparability with existing methods we used two publicly available candidates datasets: (1) PPRforNED - Pershina at el. \shortcite{pershina2015personalized};
	(2) YAGO - Hoffart at el. \shortcite{hoffart2011robust}.
	
	\paragraph{Baselines}
	As a baseline we took the standard Most Probable Sense (MPS) prediction, which selects the entity that was seen most with the given mention during training.
	We also compare to the following papers - Francis-Landau et al. \shortcite{francis2016capturing},  Yamada at el. \shortcite{yamada2016joint}, and Chisholm et al. \shortcite{chisholm2015entity}, as they are all strong local approaches and a good source for comparison.

	\paragraph{Results}
	Table \ref{tab:training} displays the micro and macro P@1 scores on CoNLL-YAGO test-b for the different training steps. We find that when using only the training set of CoNLL-YAGO our model is under-trained and that the domain adaptation significant boosts performance. We find that incorporating extra statistical and string features yields a small extra improvement in performance.
	
	The final micro and macro P@1 scores on CoNLL-YAGO test-b are displayed in table \ref{tab:conll}. On this dataset our model achieves comparable results, however it does not outperform the state-of-the-art, probably because of the relatively small training set and our reliance on domain adaptation.
	
	\begin{table}[ht]
		\begin{center}
			\begin{tabular}{|p{3.5cm}| p{1.3cm} p{1.3cm}|}
				\hline \multicolumn{3}{|c|}{CoNLL-YAGO test-b - Training Steps Eval} \\
				\hline \textbf{Model} & \textbf{Micro P@1} & \textbf{Macro P@1} \\ 
				\hline \multicolumn{3}{|c|}{PPRforNED} \\
				\hline CoNLL training set    & $82$  & $82$ \\
				+ domain adaptation    & $86.6$  & $87.7$ \\
				+ GBRT    & $87.3$  & $88.6$ \\
				\hline \multicolumn{3}{|c|}{Yago} \\
				\hline CoNLL training set    & $74.8$  & $73.5$ \\
				+ domain adaptation    & $83.6$  & $85.1$ \\
				+ GBRT    & $83.3$  & $86.3$ \\
				\hline
			\end{tabular}
		\end{center}
		\caption{\label{tab:training} Evaluation of training steps on CoNLL-YAGO.}
	\end{table}
	
	\begin{table}[ht]
		\begin{center}
			\begin{tabular}{|p{3.7cm}| p{1.2cm} p{1.2cm}|}
				\hline \multicolumn{3}{|c|}{CoNLL-YAGO test-b (Local methods)} \\
				\hline \textbf{Model} & \textbf{Micro P@1} & \textbf{Macro P@1} \\ 
				\hline \multicolumn{3}{|c|}{PPRforNED} \\
				\hline Our ARNN + GBRT    & $87.3$  & $88.6$ \\
				Yamada (2016) local               & $90.9$  & $92.4$ \\
				\hline Yamada (2016) global               & $93.1$  & $92.6$ \\
				\hline \multicolumn{3}{|c|}{Yago} \\
				\hline Our ARNN + GBRT    & $83.3$  & $86.3$ \\
				Yamada (2016) local               & $87.2$  & $89.6$ \\
				Francis-Landau (2016)             & $85.5$  & - \\
				Chisholm (2015) local             & $86.1$  & - \\
				\hline Yamada (2016) global               & $91.5$  & $90.9$ \\
				Chisholm (2015) global             & $88.7$  & - \\
				\hline
			\end{tabular}
		\end{center}
		\caption{\label{tab:conll} Evaluation on CoNLL-YAGO.}
	\end{table}
	
	\subsection{Effects of initialized embeddings and corrupt-sampling schemes}
	\label{experiments:effect}
	
	We performed a study of the effects of using pre-initialized embeddings for our model, and of using either All-Entity or Near-Misses corrupt-sampling. The evaluation was done on a $10\%$ sample of the evaluation set of the WikilinksNED corpus and can be seen in Table \ref{tab:c}. 
	
	We have found that using pre-initialized embeddings results in significant performance gains, due to the better starting point. We have also found that using Near-Misses, our model achieves significantly improved performance. We attribute this difference to the more efficient nature of training with near misses. Both these results were found to be statistically significant.

	\begin{table}[ht]
		\begin{center}
			\begin{tabular}{|c| p{1.5cm}|}
				\hline \multicolumn{2}{|c|}{Wikilinks Evaluation-Set} \\
				\hline \bf Model & \bf Micro     accuracy  \\ \hline
				\bf Near-misses, with init. &  $\bf 72.5$ \\ 
				Near-misses, random init. & $67.2$ \\ 
				All-Entity, with init. & $70$ \\ 
				All-Entity, random init. & $67.1$ \\ 
				\hline
			\end{tabular}
		\end{center}
		\caption{\label{tab:c} Corrupt-sampling and Initialization}
	\end{table}

	\section{Error Analysis}
	
	We randomly sampled and manually analyzed $200$ cases of prediction errors made by our model. This set was obtained from WikilinksNED's validation set that was not used for training. 
	
	Working with crowd-sourced data, we expected some errors to result from noise in the ground truths themselves. Indeed, we found that $19.5$\% (39/200) of the errors were not false, out of which $5\%$ (2) where wrong labels, $33\%$ (13) were predictions with an equivalent meaning as the correct entity, and in $61.5\%$ (24) our model suggested a more convincing solution than the original author by using specific hints from the context. In this manner, the  mention \textit{'Supreme leader'} , which was contextually associated to the Iranian leader Ali Khamenei, was linked by our model with \textit{'supreme leader of Iran'} while the "correct" tag was the general \textit{'supreme leader'} entity.
	
	In addition, $15.5\%$ (31/200) were cases where a Wikipedia disambiguation-page was either the correct or predicted entity ($2.5\%$ and $14\%$, respectively). We considered the rest of the 130 errors as true semantic errors, and analyzed them in-depth.
	
	\begin{table}[ht]
		\begin{center}
			\begin{tabular}{|p{3.5cm}| ll |}
				\hline \bf Error type 		& \bf Fraction  	&\\ 
				\hline \multicolumn{3}{|c|}{False errors} \\
				\hline Not errors 			& $19.5\%$ 	& $(39/200)$  \\ 
				- Annotation error 			& $5\%$    	& $(2/39)$ \\ 
				- Better suggestion			& $61.5\%$ 	&$(24/39)$ \\
				- Equivalent entities		& $ 33\%$ 	&$(13/39)$ \\ 
				Disambiguation page			& $15.5\%$   	&$(31/200)$ \\ 
				\hline \multicolumn{3}{|c|}{True semantic errors} \\
				\hline	Too specific/general  	& $31.5\%$ 	&$(41/130)$ \\ 
				'almost correct' errors		& $26\%$ 	&$(34/130)$ \\ 
				insufficient training		& $21.5\%$ &$(28/130)$ \\
				\hline
			\end{tabular}
		\end{center}
		\caption{\label{tab:d} Error distribution in $200$ samples. Categories of true errors are not fully distinct.}
	\end{table}
	
	First, we noticed that in $31.5$\% of the true errors (41/130) our model selected an entity that can be understood as a specific ($6.5$\%) or general ($25$\%) realization of the correct solution. For example, instead of predicting \textit{'Aroma of wine'} for a text on the scent and flavor of Turkish wine, the model assigned the mention \textit{'Aroma'} with the general \textit{'Odor'} entity. We observed that in $26$\% (34/130) of the error cases, the predicted entity had a very strong semantic relationship to the correct entity. A closer look discovered two prominent types of 'almost correct' errors occurred repeatedly in the data. The first was a film/book/theater type of error ($8.4$\%), where the actual and the predicted entities were a different display of the same narrative. Even though having different jargon and producers, those fields share extremely similar content, which may explain why they tend to be frequently confused by the algorithm. A third (4/14) of those cases were tagged as truly ambiguous even for human reader. The second prominent type of 'almost correct' errors where differentiating between adjectives that are used to describe properties of a nation. Particularity, mentions such as \textit{'Germanic'}, \textit{'Chinese'} and \textit{'Dutch'} were falsely assigned to entities that describe language instead of people, and vice versa. We observed this type of mistake in $8.4$\% of the errors (11/130).
	
	Another interesting type of errors where in cases where the correct entity had insufficient training. We defined insufficient training errors as errors where the correct entity appeared less than 10 times in the training data. We saw that the model followed the MPS in $75$\% of these cases, showing that our model tends to follow the baseline in such cases. Further, the amount of generalization error in insufficient-training conditions was also significant ($35.7\%$), as our model tended to select more general entities.
	
	\section{Conclusions}
	Our results indicate that the expressibility of attention-RNNs indeed allows us to extract useful features from noisy context, when sufficient amounts of training examples are available. This allows our model to significantly out-perform existing state-of-the-art models. We find that both using pre-initialized embedding vocabularies, and the corrupt-sampling method employed are very important for properly training our model.
	
	However, the gap between results of all systems tested on both CoNLL-YAGO and WikilinksNED indicates that mentions with noisy context are indeed a challenging test. We believe this to be an important real-world scenario, that represents a distinct test-case that fills a gap between existing news-based datasets and the much noisier Twitter data \cite{ritter2011Named} that has received increasing attention. We find recurrent neural models are a promising direction for this task. 
	
	Finally, our error analysis shows a number of possible improvements that should be addressed. Since we use the training set for candidate generation, non-nonsensical candidates (i.e. disambiguation pages) cause our model to err and should be removed from the candidate set. In addition, we observe that lack of sufficient training for long-tail entities is still a problem, even when a large training set is available. We believe this, and some subtle semantic cases (book/movie) can be at least partially addressed by considering semantic properties of entities, such as types and categories. We intend to address these issues in future work.
	
	\bibliographystyle{acl_natbib}
	\bibliography{_our_submission}
	
\end{document}